\documentclass[conference]{IEEEtran}
\usepackage{times}

\usepackage[numbers]{natbib}
\usepackage{multicol}
\usepackage[bookmarks=true]{hyperref}

\usepackage{graphicx}  
\usepackage{amsmath}
\usepackage{amssymb}
\usepackage{bbm}
\usepackage[ruled,vlined]{algorithm2e}
\usepackage{color}
\usepackage{xcolor}
\usepackage{subfig}
\usepackage{float}

\newcommand{\rT}{{\mathrm{T}}}

\newcommand{\E}{{\mathbb{E}}}

\newcommand{\vc}{{\bf c}}
\newcommand{\vx}{{\bf x}}
\newcommand{\vy}{{\bf y}}
\newcommand{\vz}{{\bf z}}

\newcommand{\vf}{{\bf f}}

\newcommand{\vu}{{\bf u}}

\newcommand{\vF}{{\bf F}}

\newcommand{\ExP}[2]{\E_{{#1}}{\left[#2\right]}}

\begin{document}

\title{Locally Weighted Regression Pseudo-Rehearsal for Online Learning of Vehicle Dynamics}

\author{Grady Williams, Brian Goldfain, James M. Rehg, and Evangelos A. Theodorou\\ Georgia Institute of Technology}



%

\maketitle

\begin{abstract}
We consider the problem of online adaptation of a neural network designed to represent vehicle dynamics. The neural network model is intended to be used by an MPC control law to autonomously control the vehicle. This problem is challenging because both the input and target distributions are non-stationary, and naive approaches to online adaptation result in catastrophic forgetting, which can in turn lead to controller failures. We present a novel online learning method, which combines the pseudo-rehearsal method with locally weighted projection regression. We demonstrate the effectiveness of the resulting Locally Weighted Projection Regression Pseudo-Rehearsal (LW-PR$^2$) method in simulation and on a large real world dataset collected with a 1/5 scale autonomous vehicle. 
\end{abstract}

\IEEEpeerreviewmaketitle

\section{Introduction}

Autonomous vehicles operating in the real world must be capable of precise control in order to satisfy safety constraints and ensure comfort. This, in turn, requires highly accurate dynamics models which can be difficult to obtain. One of the most promising approaches is to learn the dynamics using function approximation methods. However, a major challenge is that the system dynamics are constantly changing, either due to environmental factors (e.g. the road surface condition), or due to internal factors that are hard to measure (e.g. weight distribution, suspension stiffness, steering slop). Therefore, models which are able to adapt to changing dynamics online are necessary.

In this work we examine the use of a neural network to model the system dynamics. Neural networks have recently been demonstrated to be effective at learning system models that can be used in high performance vehicle controllers \cite{williams2018information}, and are seeing increasing usage in the robotics control literature \cite{nagabandi2018neural, lenz2015deepmpc, finn2017deep, clavera2018learning}. Unfortunately, adapting neural networks online is a notoriously difficult problem. The key issue that must be overcome is catastrophic forgetting \cite{mccloskey1989catastrophic, ratcliff1990connectionist}, which is the tendency for neural networks to forget old data when fed new data from a different distribution. This is especially problematic in the case of robotics control, where catastrophic forgetting in the system model can lead to a controller failure.

In order to overcome the catastrophic forgetting problem in neural networks, we make use of a class of modeling techniques which are naturally immune to catastrophic forgetting: locally weighted linear regression \cite{atkeson1997locally}. Locally weighted linear regression methods work by building a global model up from a set of many small local linear models. Each model is equipped with a receptive field, which determines how much the model should respond to a given input. The output of the global model is then computed as a weighted average of all the local model outputs \cite{schaal1998constructive, vijayakumar2005incremental}. Since a given training pair only affects a highly localized region of the state space, locally weighted regression methods can safely make incremental updates. One of the most mature local modelling methods is locally weighted projection regression (LWPR) \cite{vijayakumar2005incremental}. LWPR has proven successful in modeling robot dynamics, even for high-dimensional systems in an online learning scenario.

Given the success of LWPR at the task of learning robot dynamics, and the ability for it to be used in an incremental online setting, one may wonder: why not use LWPR instead of neural networks for learning system dynamics? The issue with utilizing locally linear methods in model predictive control is the computational cost. In the past, locally weighted regression methods have been limited to inverse control, which only requires a single prediction per timestep, or offline trajectory/policy optimization \cite{atkeson1997locally, mitrovic2008adaptive}. In cases where they have been used in MPC \cite{williams2015gpu}, the model had to be severely restrained in order to control the number of local models generated. The issue is that for local linear methods to achieve high accuracy, usually thousands or even tens of thousands of local models are required to be effective. This requires an order of magnitude more floating point operations than a neural network to get comparable prediction accuracy. Thus, utilizing locally linear methods instead of neural networks inside of an MPC method would come at a significant opportunity cost, if it were even possible to run in real-time.

Instead of directly utilizing LWPR in the MPC controller, our approach will be to maintain both a neural network model and an LWPR model. The neural network model is used by an MPC controller and it is updated online using data recently collected from the system, but it is also regularized using pseudo-samples generated by an LWPR model. The LWPR model is updated in an incremental online manner, in order to ensure that the artificially generated data matches the current target distribution.

\section{Related Work on Neural Network Adaptation}

Some of the earliest methods proposed for mitigating catastrophic forgetting in neural networks are rehearsal and pseudo-rehearsal \cite{ratcliff1990connectionist, robins1993catastrophic, robins1995catastrophic, robins2004sequential, french1997using}. In rehearsal methods, the original training data is retained and used alongside the new data in order to update the model. Pseudo-rehearsal methods do not retain old data, but instead they randomly create input vectors (pseudo-inputs) that are then fed through the current network in order to produce a corresponding output point (pseudo-output). The resulting artificially generated sample (pseudo-sample) can then be used for training the network alongside newly received data. The idea is that, by using the pseudo-samples alongside real data, the network can be encouraged to learn the new data without forgetting the current mapping. Recently, there has been success using rehearsal \cite{rebuffi2017icarl} and pseudo-rehearsal based methods for vision tasks \cite{shin2017continual, atkinson2018pseudo, mellado2017pseudorehearsal, kemker2017fearnet}. In these methods the primary challenge that must be overcome is either storing previous data samples (in rehearsal methods) or randomly generating realistic inputs (for pseudo-rehearsal methods).

In the case of learning vehicle dynamics, generating pseudo-inputs is relatively easy due to the low dimensional state-space representation of a vehicle. Instead, there is another challenge that must be overcome that cannot be handled by the usual rehearsal or pseudo-rehearsal techniques: \emph{both} the input and target distributions are non-stationary. This means that sometimes we need to learn new data while retaining old data, which is the case when encountering a novel region of the state space. But, other times, we need to learn new data which overwrites old data, this is the case if we are in a familiar region of the state-space but the target distribution has changed. The approach we develop to handle this problem is best thought of as a type of pseudo-rehearsal method, with the key innovation being the use of an incrementally updated LWPR model to produce the pseudo-outputs. We also make use of a constrained gradient descent update rule in order to prevent large errors on new training data from overwhelming the training signal on previously seen data.

Besides rehearsal and pseudo-rehearsal methods, there are a variety of methods for updating neural networks which mitigate catastrophic forgetting by controlling how far the parameters of the model can move away from the current model. For instance, this is the approach taken in \cite{wan2006parameter, kirkpatrick2017overcoming, li2018learning}. However, as in the case of rehearsal and pseudo-rehearsal, it is not clear how controlling changes in the weights works when the target distribution is variable, since in that case the network weights corresponding to previously learned data will need to be changed as well. Another promising approach to online adaptation for neural networks that has recently been explored is meta-learning \cite{clavera2018learning}. However the meta-learning approach does not have an explicit mechanism to combat catastrophic forgetting, and it is currently unclear how to perform the meta-training in order to ensure that catastrophic forgetting cannot occur.

\section{Problem Formulation}

Consider an autonomous vehicle operating at some task, while performing the task the vehicle encounters system states, denoted $\vz$, and executes controls, denoted $\vu$. Our goal is to update the model of the vehicle dynamics, which is defined using the following discrete time dynamical system:
\begin{equation}
    \vz_{t+1} = \vz_t + \vF(\vz_t, \vu_t; \theta)\Delta t
\end{equation}
Where $\theta$ denotes the parameters of the model, in our case these are the weights of a neural network. The system states for the vehicle are position, heading, roll, velocities, and heading rate, and the control inputs are steering and throttle commands. In the case of a ground vehicle, the position and heading updates are kinematically trivial, so we can re-write the dynamics as:
\begin{align}
\vz_{t+1}^k &= \vz_t^k + k(\vz_t)\Delta t \\
\vz_{t+1}^d &= \vz_t^d + \vf(\vz_t^d, \vu_t; \theta)\Delta t
\end{align}
Here $\vz^k$ denotes kinematic states, which are position and heading, and $\vz^d$ denotes dynamics states, which are roll, body-frame longitudinal and lateral velocity, and heading rate. The motion update for the kinematic states is trivial, so we need only focus on learning $\vf(\vz_t^d, \vu_t; \theta)$. Now, we define the following variables:
\begin{equation}
\vx = 
\begin{pmatrix}
\vz_t^d \\
\vu_t
\end{pmatrix},
~~~~
\vy = \frac{\vz_{t+1}^d - \vz_t^d}{\Delta t}
\end{equation}
as the inputs and targets for our learning algorithm. Now, as the vehicle moves about in the world, it encounters states and controls according to some probability distribution:
\begin{equation}
\text{Local Operating Distribution}: ~~ \vx \sim \mathcal{P}_L(\mathcal{X}).
\end{equation}
The distribution $\mathcal{P}_L$ is called the local operating distribution, and it is highly task dependent. In addition to the local operating distribution, we assume that there is a system identification dataset, which contains data consisting of all the various maneuvers that the vehicle needs to learn in order drive competently. The system identification dataset consists of samples drawn from another distribution:
\begin{equation}
\text{System Identification Distribution}: ~~ \vx \sim \mathcal{P}_{ID}(\mathcal{X}).
\end{equation}
which we denote as $\mathcal{P}_{ID}$, and refer to as the system identification distribution. Note that this distribution is constant, but the mapping which takes input points drawn from this distribution to the corresponding dynamics output is not. Our goal is to incrementally update a neural network describing the system dynamics. However, we must be sure that by updating the model we do not forget any of the system modes contained in $\mathcal{P}_{ID}$. 

First consider a simple approach to performing online model adaptation based on standard stochastic gradient descent (SGD). Suppose that we have access to streaming data, and that we maintain a set of recently encountered input and output pairs. By randomly drawing pairs from this set, we can get independent and identically distributed (I.I.D.) samples from the local operating distribution. The standard SGD rule updates the parameters as follows:
\begin{equation}
\theta_{i+1} = \theta_i - \gamma \nabla_{\theta_i} \| \vy - \vf(\vx; \theta_i) \|^2
\end{equation}
since training pairs are drawn from the local operating distribution, this update will improve the model's performance for the inputs drawn from that distribution. Mathematically, this means that we are optimizing for the objective: 
\begin{equation}
\ExP{\vx \sim \mathcal{P}_L}{\| \vy - \vf(\vx; \theta) \|^2}
\end{equation}
This is \emph{not} what we want. The issue with this is that the local operating distribution may not contain all the maneuvers that the vehicle needs to operate effectively. A typical example in the case of autonomous driving is highway driving: a vehicle operating on a highway only needs to maintain a constant velocity and make slight turns the vast majority of the time, if the model is updated with inputs purely drawn from a highway driving dataset, there is no guarantee that the model will remember the basic maneuvers necessary for other types of driving. This problem, known as \emph{catastrophic forgetting}, is a well known deficiency of neural networks, and it is especially problematic when the model being updated is being used to control the vehicle.

If we had access to I.I.D. samples from the system identification dataset, we could instead use an SGD update law that jointly learns the target mapping for inputs drawn both from the system identification dataset and the local operating distribution. For instance, the following update law achieves this:
\begin{align}
\label{Equation:GradientComputation}
\theta_{i+1} &= \theta_i - \gamma \left( G_L(\theta_i) + G_{ID}(\theta_i) \right) \\
G_L &= \nabla_{\theta_i} \| \vy_L - \vf(\vx_L; \theta_i) \|^2 \\ 
G_{ID} &= \nabla_{\theta_i} \| \vy_{ID} - \vf(\vx_{ID}; \theta_i) \|^2\\
(\vx_L, \vy_L) &\sim \mathcal{P}_L, ~~~ (\vx_{ID}, \vy_{ID}) \sim \mathcal{P}_{ID}
\end{align}
This update balances optimizing the model on the local operating distribution and the system identification dataset, and it is the basic idea behind traditional rehearsal and pseudo-rehearsal techniques. In the case of classification, this type of update is effective at preventing catastrophic forgetting. But, in the regression setting, even this type of update could be problematic since the magnitude of the error incurred by the network can vary greatly depending on the region of state-space the system is in. If the error incurred by the local operating distribution is very high it can overwhelm the error signal from the system identification part of the data, which can still lead to the vehicle forgetting basic maneuvers.  

Instead, we want to ensure that the model cannot forget the system identification dataset, in this context ``forgetting'' means intentionally degrading the performance of the model on input data drawn from $\mathcal{P}_{ID}$. One way to enforce that is to ensure that update steps always move in the direction of the local minima for input data drawn from the system identification distribution. This constraint can be enforced by ensuring that the cosine of the angle between the update direction and the gradient computed from the system identification data is always positive, and it can be achieved with the following update law:
\begin{align}
\label{Equation:UpdateRule}
&\theta_{i+1} = \theta_i - \gamma  \left( \alpha G_L(\theta_i) + G_{ID}(\theta_i) \right) \\
&\alpha = \max_{a \in [0,1]} ~~ s.t ~~ \langle{a G_L(\theta_i) + G_{ID}(\theta_i) , G_{ID}(\theta_i) }\rangle \ge 0
\end{align}
This update law still balances the objective of simultaneously optimizing for local operating distribution and system identification distribution. However, it constrains the combined gradient to always point in the same direction as the gradient computed from system identification data.

The problem with implementing the update rule defined by Eq. \eqref{Equation:UpdateRule} is that, in an online setting, we only have access to data generated from the local operating distribution. Additionally, since the target mapping is changing, we cannot simply re-draw samples from the original system identification dataset or generate pseudo-outputs by running random inputs through the current model like in standard rehearsal and pseudo-rehearsal methods.

\section{Locally Weighted Projection Regression Pseudo-Rehearsal}

Our goal is to approximately implement the constrained gradient update defined in Eq. \eqref{Equation:UpdateRule}. Our strategy will be to use artificially generated pseudo-training points to enforce the constraint, with the additional requirement that the pseudo-training points must somehow match the changing target distribution. This means that artificially generating training points requires two steps:
\begin{enumerate}
    \item A method for generating artificial input points that are I.I.D. samples from $\mathcal{P}_{ID}(\mathcal{X})$.
    \item A method for computing the corresponding target, $\vy$, for an artificially generated input point. This should be a function approximator that is capable of online adaptation, since the target mapping $\vy$ is actively changing.
\end{enumerate}

Given these requirements, it appears as though we cannot go anywhere, since the preceding discussion can roughly be summarized as: in order to perform online adaptation we first need a method that can do online adaptation in order to constrain the stochastic gradient descent update. The reason why this is not the case, is that the speed requirements on the model used for artificial data generation are much less strenuous than the speed requirements for a model used in model predictive control. 

Suppose that we are receiving inputs points at a rate of 40 Hz, a reasonable number for a robotic system, then in order to produce an equal number of artificial points for regularization our model needs to be able to produce 40 predictions/second. In contrast, for our sampling based MPC controller we require the model to produce on the order of millions of predictions per second. Even less computationally intensive MPC algorithms (iLQG for instance) require on the order of tens of thousands of predictions (plus derivative computations) per second. So, even for relatively inexpensive MPC algorithms the computational demands on the model used for artificial data generation are 3 orders of magnitude less than the model used for MPC, and 5 orders of magnitude less than for our sampling based controller. This means that we can use more computationally demanding models that are specifically suited to online adaptation in order to generate the artificial data, such as LWPR.

\subsection{Algorithm Description}

\begin{figure*}[htb!]
    \centering
    \includegraphics[width=\textwidth]{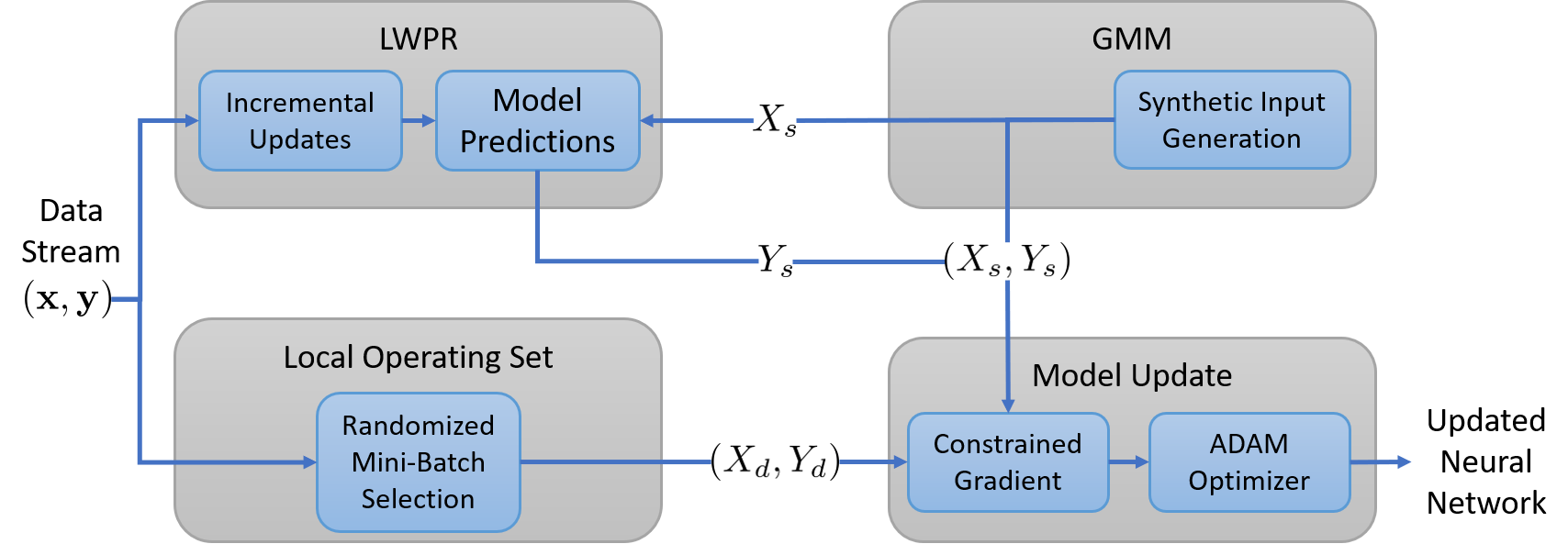}
    \caption{LW-PR$^2$ Algorithm. The GMM produces synthetic input points which are combined with predictions from LWPR to create synthetic training pairs. These are combined with randomized mini-batches created from recently collected data in order to compute the constrained gradient update.}
    \label{fig:LW-PR2-Diagram}
\end{figure*}

Our approach will be to train an LWPR model, which will be updated online, in order to compute the target mapping for artificially generated input points. For the generation of the input points, a gaussian mixture model (GMM) is used. The GMM is trained offline and kept static, which reflects the fact that the input distribution defined by the system identification dataset does not change. The artificial input/output pairs are then used to compute a synthetic gradient, which is used to regularize the online stochastic gradient descent. The algorithm consists of four sub-modules, which we now describe in detail. The overall flow of the algorithm is shown in Fig. \ref{fig:LW-PR2-Diagram}. 

\subsubsection{Gaussian Mixture Model}

The purpose of the GMM is to generate synthetic input points consistent with the system identification dataset. We denote a mini-batch of synthetic input points as $X_s$. The GMM uses diagonal covariance matrices, and is trained using standard expectation maximization. We use the Bayesian Information Criterion (BIC) in order select the number of gaussian models used. After the initial training the GMM is not modified again. This is because the input distribution for the system identification dataset should be carefully curated in order to ensure that it contains a balance of all necessary maneuvers.

\subsubsection{LWPR Module}

The LWPR module takes in the synthetic input points generated by the Gaussian mixture model, and then runs those input points forward through the LWPR model in order to produce synthetic output points. If we let $\vy_i$ and $\vc_i$ be the mean and center of the $i_{th}$ local model and let $D_i$ be the distance metric which defines the receptive field for the $i_{th}$ model, then LWPR computes the global prediction as:
\begin{equation}
\vy(\vx) = \sum_{i=1}^L w_i \cdot \vy_i(\vx - \vc_i)
\end{equation}
Where the weight governing the response of each local model is: 
\begin{equation}
w_i = \frac{\exp\left(-\frac{1}{2}(\vx - \vc_i)^\rT D_i (\vx - \vc_i)\right)}{\sum_{j=1}^L \exp\left(-\frac{1}{2}(\vx - \vc_j)^\rT D_j (\vx - \vc_j)\right)}
\end{equation}
Since the response of a given model to an input decays exponentially fast, model updates have only a negligible impact on models with centers far from the current input point. This is the feature that makes LWPR largely immune to catastrophic forgetting, so it can be safely updated online.

The mini-batch output of the LWPR module is denoted $Y_s$. The LWPR model is initially trained over several epochs on the system identification dataset using the standard LWPR update rule. Online, the local linear models making up the LWPR model are continually updated. We train one LWPR model for each different output dimension (roll rate, longitudenal acceleration, lateral acceleration, and heading acceleration). The LWPR module is implemented using \cite{klanke2008library}.

\subsubsection{Local Operating Set}

The local operating set consists of the last several seconds of training points received from the stream of data generated by the system. In our implementation this set contains between 500 and 1000 (10 - 20 seconds) of data. Out of this set of data, randomized mini-batches are created (denoted as $(X_d, Y_d)$) and then fed into the model updater.

\subsubsection{Model Update}

The last module in the LW-PR$^2$ algorithm is the computation of the constrained gradient and the actual model update. First, the gradient for the local operating distribution, $G_L$, is computed using the mini-batch received from the local operating distribution, and then the gradient for the system identification dataset, $G_{ID}$ is computed using the artificial mini-batch received from the GMM and LWPR modules. The constrained gradient is computed via equations \eqref{Equation:GradientComputation} - \eqref{Equation:UpdateRule}. After the gradient is computed we use the ADAM optimizer \cite{kingma2014adam} to perform the update step.

\subsection{Neural Network Initialization}

Before the neural network can be updated, an initial model needs to be trained on the original system identification dataset. One option is to initialize the model using standard stochastic gradient descent (i.e. without taking into account the other modules), but we have found it is more effective to jointly train the initial model with the LWPR model and GMM model. This means that the actual system identification dataset takes the place of the local operating set, but synthetic data is still generated by the GMM and LWPR modules, which is used to compute the constrained gradient. We have observed that training the initial model in this manner has a negligible effect on the performance of the initial trained model, but helps with the adaptation.

\section{Results}

We tested our locally weighted projection regression pseudo-rehearsal (LW-PR$^2$) approach using four different sets of experiments. All of the data in these experiments are collected using vehicles based on the design from \cite{Autorally} or the simulation tools from \cite{Autorally}. These vehicles are called ``AutoRally'' platforms, and they are highly robust 1/5 scale autonomous vehicles which have realistic dynamics compared to a full size platform. 

Our first experiment tests the algorithm's ability to prevent catastrophic forgetting, using a dataset designed specifically to induce catastrophic forgetting on naive adaptation methods. The second experiment tests the method's ability to adapt to drastic changes in the system dynamics using a driving dataset collected on a muddy surface. The third experiment tests how effective the updated model is when used as part of an MPC algorithm. Lastly, the fourth experiment consists of running the model adaptation during a full day of testing with the 1/5 scale autonomous driving system, and measures how well the algorithm works in a practical setting. Throughout these experiments there are 3 different types of experimental modes that are run:
\begin{enumerate}
    \item An {\bf{offline}} test is a test where the model is not allowed to adapt during the experiment.
    \item An {\bf{online}} test is a test where the model is allowed to adapt during the experiment, but the adapted model is not used to control the vehicle. 
    \item An {\bf{active}} test is a test where the model is allowed to adapt during the experiment, and the adapted model is used to control the vehicle.
\end{enumerate}
Recall that in an online training scenario, there is not an explicit training, validation, and test set. Instead, as each training pair is received, we compute the current error on that training pair, and we then record the result. After the error has been computed and recorded, the training pair is fed into the model updater. We compare our method against the base model (no adaptation), and the base model adapted with standard stochastic gradient descent. We also record the performance of the LWPR model used to generate synthetic training inputs. 

It is important to realize that the LWPR model we use for generating synthetic data would not be feasible for use in a real-time control loop. Tables \ref{Table:lwprComputation} and \ref{Table:neuralNetComputation} detail the computational requirements of the neural network and LWPR respectively. For these calculations, we assume that a dot product operation takes $2N-1$ floating point operations (FLOPs) for vectors of dimension $N$, and that a matrix-vector multiplication takes $2MN - M$ FLOPs where the matrix has dimension $M \times N$. We also assume that any non-linear function ($\exp, \tanh, (\cdot)^2$) takes a FLOP. For LWPR it can be difficult to predict the throughput required since the number of active local models can vary greatly, so we compute a lower bound based only on how many local model activations must be computed. Computing a local model activation requires first subtracting the mean for the local model from the current input point ($6$ FLOPs), then individually squaring each result ($6$ FLOPs), then computing the dot product between the result and the receptive field weight ($2\cdot6 - 1$ FLOPs), and then computing the negative exponential of the result ($2$ FLOPs). This results in a total of 25 floating point operations for each local model, plus the additional computations required to actually compute the weighted average. The neural network simply consists of matrix-vector multiplies, the additions of the bias, and $\tanh$ non-linearities. Note also, that LWPR works best when a separate model is used for each output dimension, whereas only one neural network is required.

\begin{table}[b!]
\caption{LWPR Computational Requirements}
\centering
\begin{tabular}{c|c|c}
Output Variable      & Receptive Fields & FLOPs/Pred   \\ \hline
Roll Rate            & $162$    & $>4,050$                 \\ \hline
Longitudinal Acc.    & $1,409$   & $>35,225$                \\ \hline
Lateral Acc.         &     $1,738$ & $>43,450$                \\ \hline
Heading Acc.         &     $2,336$    & $>58,400$                 \\ \hline
Total & $5,645$    & $>141,125$
\end{tabular}
\label{Table:lwprComputation}
\end{table}

\begin{table}[b!]
\caption{Neural Network Computational Requirements}
\centering
\begin{tabular}{c|c|c}
Layer Transition    & Input - Output Neurons & FLOPs/Pred   \\ \hline
Input - Hidden 1 & 6 - 32 & 416  \\ \hline
Hidden 1 - Hidden 2 & 32- 32 & 2,080 \\ \hline
Hidden 2 - Output & 32 - 4 & 256 \\ \hline
Total &  6-32-32-4 & 2,688
\end{tabular}
\label{Table:neuralNetComputation}
\end{table}

The key takeaway from Tables \ref{Table:lwprComputation} and \ref{Table:neuralNetComputation} is that making predictions with the neural network is two orders of magnitude cheaper than making predictions with LWPR models. Since MPC controllers need to make tens of thousands or millions of predictions per second, this is important. For instance, our MPC controller performs 6 million dynamics predictions every second. If we used the LWPR model that we have trained we would need to achieve a throughput of at least 847 GFLOP/S to run in real time. Although this number is technically achievable for modern graphics cards on dense matrix multiplication benchmarks (the Nvidia GTX 1050 Ti in our AutoRally platform has a peak measured performance of 1.8 TFLOPS), it is not currently possible for algorithms with more complicated memory usage, control flow and synchronization requirements - such as forward propagating an LWPR model. In contrast, the synthetic data generation only requires on the order of tens or hundreds of predictions per second, which is easily manageable on any reasonably capable modern processor.

\subsection{Catastrophic Interference Test}

In this experiment we test the ability of LW-PR$^2$ to improve online modeling performance while simultaneously ``remembering" other parts of the system identification dataset. These experiments utilize two datasets, which we selected from the publicly available dataset accompanying \cite{williams2018information}. The first dataset we call the \emph{online training dataset} and the second is called the \emph{offline validation dataset}. These datasets were collected on the same day, so the environmental differences are minimal\footnote{The offline validation dataset was collected approximately 30 minutes before the online training dataset}. The online training dataset consists of 100 laps (approximately 27 minutes) of slow speed driving around a roughly elliptical track in the clockwise direction. This is an extremely monotonous dataset, as the robot mostly follows the same line over the 100 laps.

Given the monotonous nature of the online training dataset, it is easy for the online adaptation to overfit to the local operating distribution and forget parts of the system identification dataset. In order to test how well the model adaptation is able to remember other system modes, we utilize the offline validation dataset. The offline validation dataset consists of the same type of low speed monotonous driving in the \emph{opposite} (counter-clockwise) direction of the online training dataset. If the model adaptation algorithm is successful at remembering the system identification dataset, then we should be able to run the adaptation on the online training dataset and see minimal degradation when testing the final adapted model on the offline validation dataset.

\begin{table}[htb!]
\caption{Online Training Dataset Errors}
\centering
\begin{tabular}{c|c|c|c|c} 
                            & Base & SGD & LW-PR$^2$ & LWPR  \\ \hline
Roll Rate ($rad./s$)        & 0.01  & 0.01& 0.01 &  0.01   \\ \hline
Longitudinal Acc. $(m/s^2)$ & 0.35 &0.33 & {\bf{0.32}} &  0.33   \\ \hline
Lateral Acc. $(m/s^2)$      & 0.69 &0.63 & {\bf{0.61}} &  0.65   \\ \hline
Heading Acc. $(rad./s^2)$   & 2.11 &{\bf{0.46}} & 0.49 &  0.53   \\ \hline
Total MSE                   & 0.79 &{\bf{0.36}} & {\bf{0.36}} &  0.38
\end{tabular}
\label{Table:OnlineDynamicsPerformance}
\end{table}
\begin{table}[htb!]
\caption{Offline Validation Dataset Errors}
\centering
\begin{tabular}{c|c|c|c|c} 
                            & Base & SGD & LW-PR$^2$ & LWPR  \\ \hline
Roll Rate ($rad./s$)        & 0.01 & 0.01& 0.01 &  {\bf{0.00}}   \\ \hline
Longitudinal Acc. $(m/s^2)$ & 0.20 &0.22 & 0.20 &  {\bf{0.17}}   \\ \hline
Lateral Acc. $(m/s^2)$      & 0.99 &1.56 & 1.10 &  {\bf{0.83}}   \\ \hline
Heading Acc. $(rad./s^2)$   & 1.47 &2.11 & 0.83 &  {\bf{0.42}}   \\ \hline
Total MSE                   & 0.67 &0.97 & 0.53 &  {\bf{}0.36}
\end{tabular}
\label{Table:OfflineValidationPerformance}
\end{table}

The testing procedure works as follows: we first test the online performance of each of the methods using the online training dataset. Then, after the online test is finished, the final adapted model is taken and an offline test (no adaptation allowed) is performed on the offline validation dataset. The results of these tests are shown in Tables \ref{Table:OnlineDynamicsPerformance} and \ref{Table:OfflineValidationPerformance}.

For the online training dataset all of the adaptive methods perform similarly, and they all significantly decrease the total mean-squared error of the model predictions compared with the base neural network model. However, when using the final adapted model from the online training dataset on the offline validation dataset, the differences between the methods become apparent. The standard SGD methods suffers the characteristic catastrophic forgetting, particularly in the heading acceleration which makes sense given the difference in the direction of travel between the two datasets. As expected, LWPR is unaffected by the change in local operating distribution, and performs better than the base network. Our LW-PR$^2$ method performs only slightly worse than LWPR and outperforms the base network. This shows that the method is not only capable of preventing catastrophic forgetting, but that it is actually able to generalize knowledge gained in one dynamics regime to another related dynamics regime.

\subsection{Modified Dynamics Test}

In this experiment we test the ability of the algorithm to adapt to highly modified vehicle dynamics. The dynamics are modified by running the vehicle on a muddy track surface. The mud changes the dynamics because it clings to the tires and reduces the friction between the vehicle and the ground. Additionally, the mud clinging to the body of the vehicle adds over 10 kg of extra weight (the normal weight is 21 kg), which has a significant effect on the vehicle's dynamics. All of the system identification data was collected on a dry surface with a mud free robot, so this is a completely novel dynamics regime for the system.
\begin{figure}[htb!]
\centering
\includegraphics[width=\columnwidth]{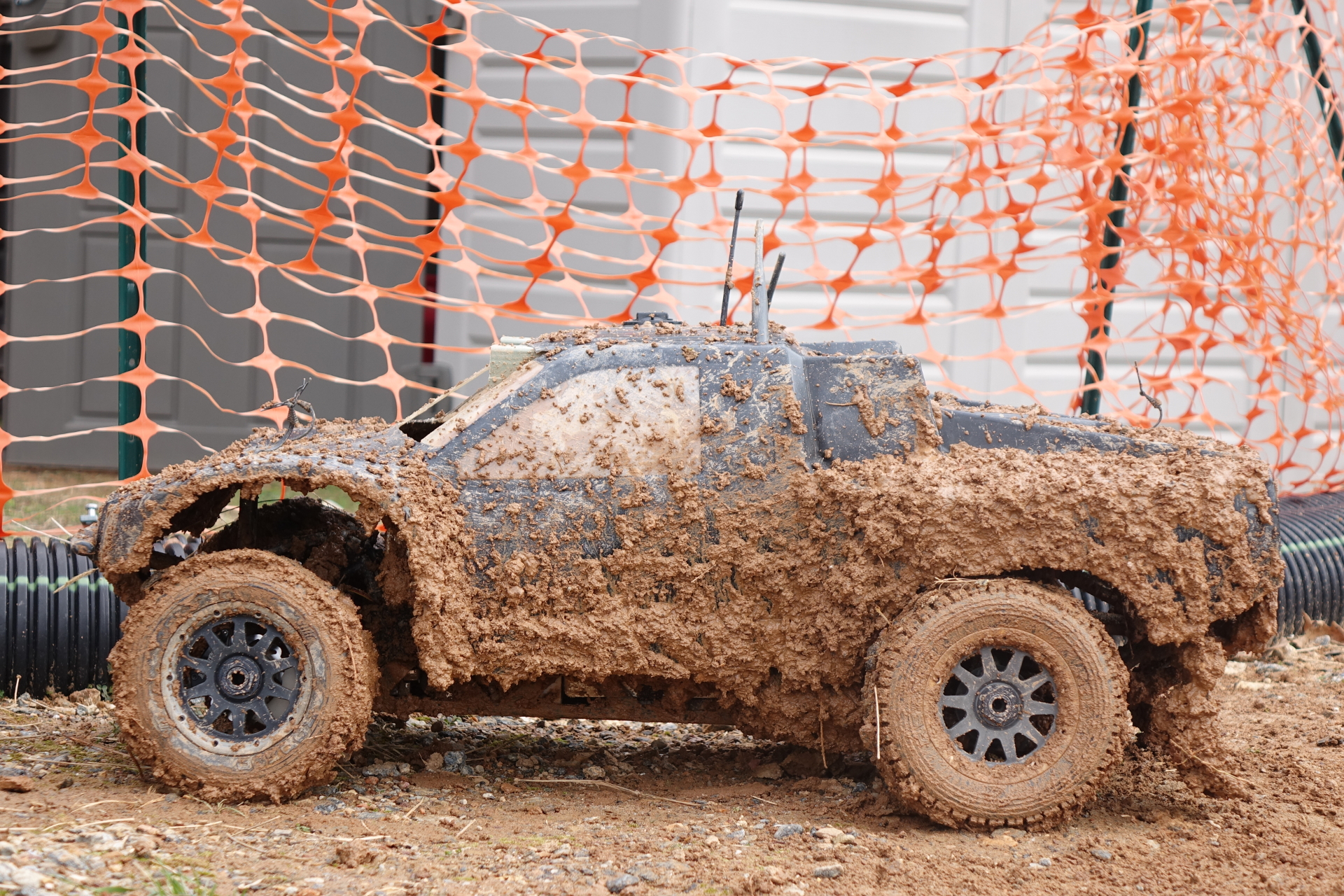}
\label{Fig:Muddy}
\caption{AutoRally vehicle after running on muddy surface. Notice the depressed rear suspension, which is caused by the extra weight from the mud accumulating on the chassis and body.}
\end{figure}

The vehicle is driven by an expert 1/5 scale RC car driver in the muddy conditions. Despite the poor driving conditions, the driver is still able to attain speeds over 50 kph and slip angles in excess of 60 degrees. This means that the dataset is challenging not only because of the changing conditions, but also because of the highly dynamic regime that the vehicle operates in. This dataset consists of slightly more than 2.5 minutes of data, which is 5 laps around our test track. The results are given in Table \ref{Table:MuddyPerformance}.
\begin{table}[htb!]
\caption{Modified Dynamics Dataset Errors}
\centering
\begin{tabular}{c|c|c|c|c} 
                            & Base & SGD & LW-PR$^2$ & LWPR  \\ \hline
Roll Rate ($rad./s$)        & 0.02& 0.02& 0.02 &  0.02   \\ \hline
Longitudinal Acc. $(m/s^2)$ & 2.42 &{\bf{1.52}} & 1.60 &  1.70   \\ \hline
Lateral Acc. $(m/s^2)$      & 1.04 &1.02 & {\bf{0.96}} &  0.98   \\ \hline
Heading Acc. $(rad./s^2)$   & 4.96 &{\bf{2.64}} & 2.74 &  3.10   \\ \hline
Total MSE                   & 2.11 &{\bf{1.29}} & 1.33 &  1.45
\end{tabular}
\label{Table:MuddyPerformance}
\end{table}

All of the incremental methods achieve a better total MSE than the un-modified base network. The LW-PR$^2$ method actually outperforms LWPR by a significant margin. The standard SGD method performs best, as it did in the catastrophic forgetting test on the online training dataset. The weakness of SGD is not that it does not fit the local operating distribution, but that it can easily forget the system identification dataset.

\subsection{Simulated Autonomous Driving Tests}

The previous two experiments tested our method on datasets where the model being produced by the model adaptation was not being used to control the vehicle. In this section, we test the algorithm in an active setting where a model predictive controller uses the updated model to control the vehicle. We use the same open source Gazebo simulation of the AutoRally vehicle from \cite{Autorally} for these experiments. The system identification dataset that we use to train the base model is the same as in the previous sections (i.e. it is based on real world data). Note that the simulation dynamics are significantly different from the real-world AutoRally dynamics, so the starting base model is highly inaccurate.

We ran three different model adaptation settings: standard SGD, LW-PR$^2$, and no adaptation. For each setting we performed trials running 10 laps around the track, and we collected 5 trials for each different setting. The vehicle is driven using the model predictive path integral controller from \cite{williams2018information}, with a desired speed set\footnote{For speeds higher than $8 m/s$ the accelerations reported by the simulator exhibited high frequency oscillations indicating numerical instability of the simulator, this prevented us from trying faster speeds using the model adaptation.} at $8 m/s$. In order to emphasize the impact of the model adaptation, we set the maximum slip angle the vehicle is allowed to achieve to be relatively low ($\le 13$ degrees). Since the vehicle slides less easily in the gazebo simulation than in the real world, this results in a very conservative controller when using the base model (which has only seen real-world data). As the model adapts, the controller should realize that it can increase speed without slipping, leading to improved performance.

The results of all of the trials are shown in Table \ref{Table:GazeboSimResults}. Using either the base model or the LW-PR$^2$ adapted model, the controller is able to successfully navigate around the track. However, when using standard SGD to update the model the controller consistently fails after completing 1 lap, typically the controller tries to take a turn too fast, which results in the vehicle rolling over.
\begin{table}[htb!]
\caption{Gazebo Simulation Results}
\centering
\begin{tabular}{c|c|c|c} 
                              & Base Network & ~~SGD~~ & LW-PR$^2$ \\ \hline
Avg. Laps Completed           &  10    &    1    &    10       \\ \hline
Avg. Trial MSE                &  1.84  &  2.49   &  {\bf{0.65}}       \\ \hline
Avg. Lap Time                 &  34.78 &   N/A   & {\bf{32.04}}       \\ 
\end{tabular}
\label{Table:GazeboSimResults}
\end{table}

\begin{figure}[htb!]
    \centering
    \includegraphics[width=0.9\columnwidth]{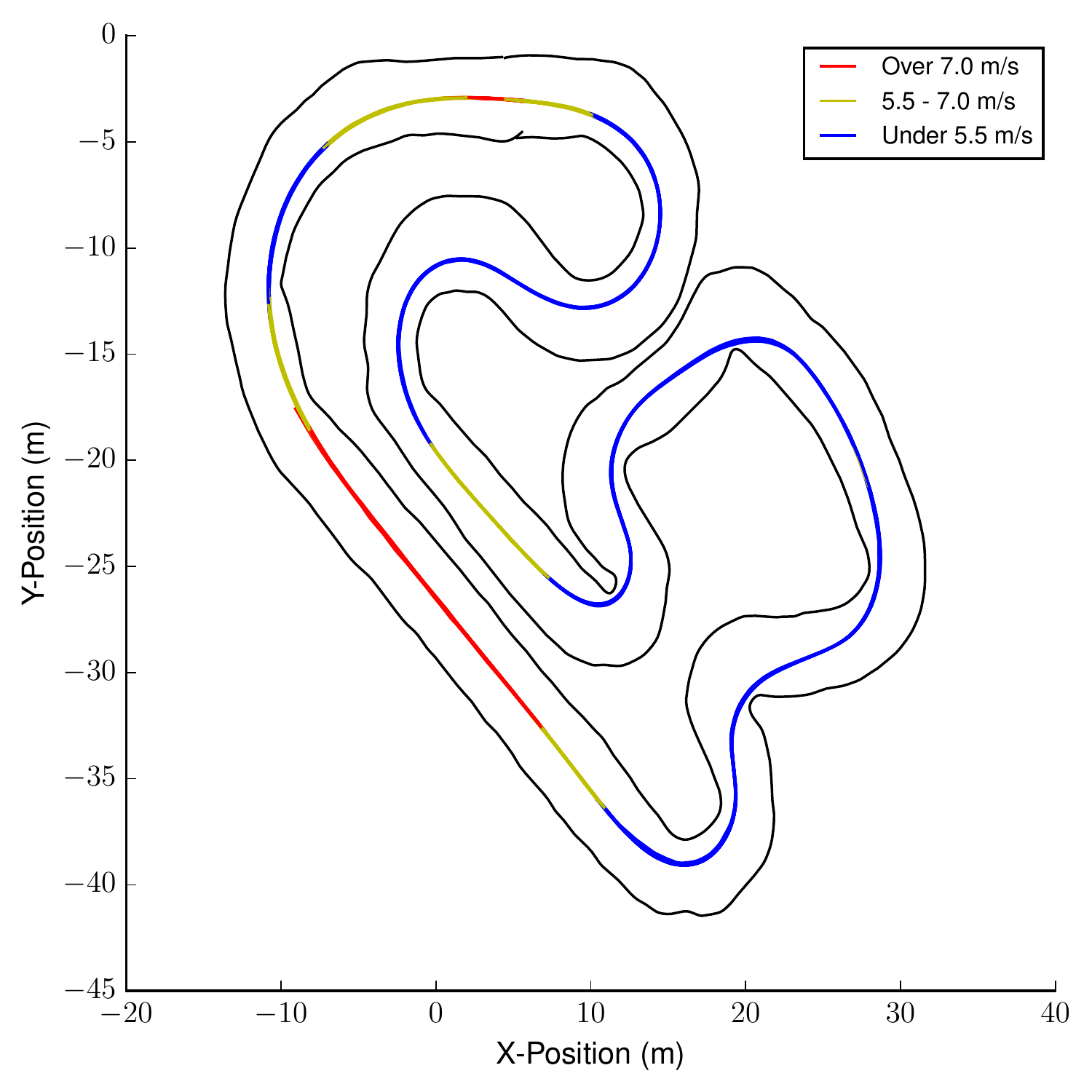}
    \includegraphics[width=0.9\columnwidth]{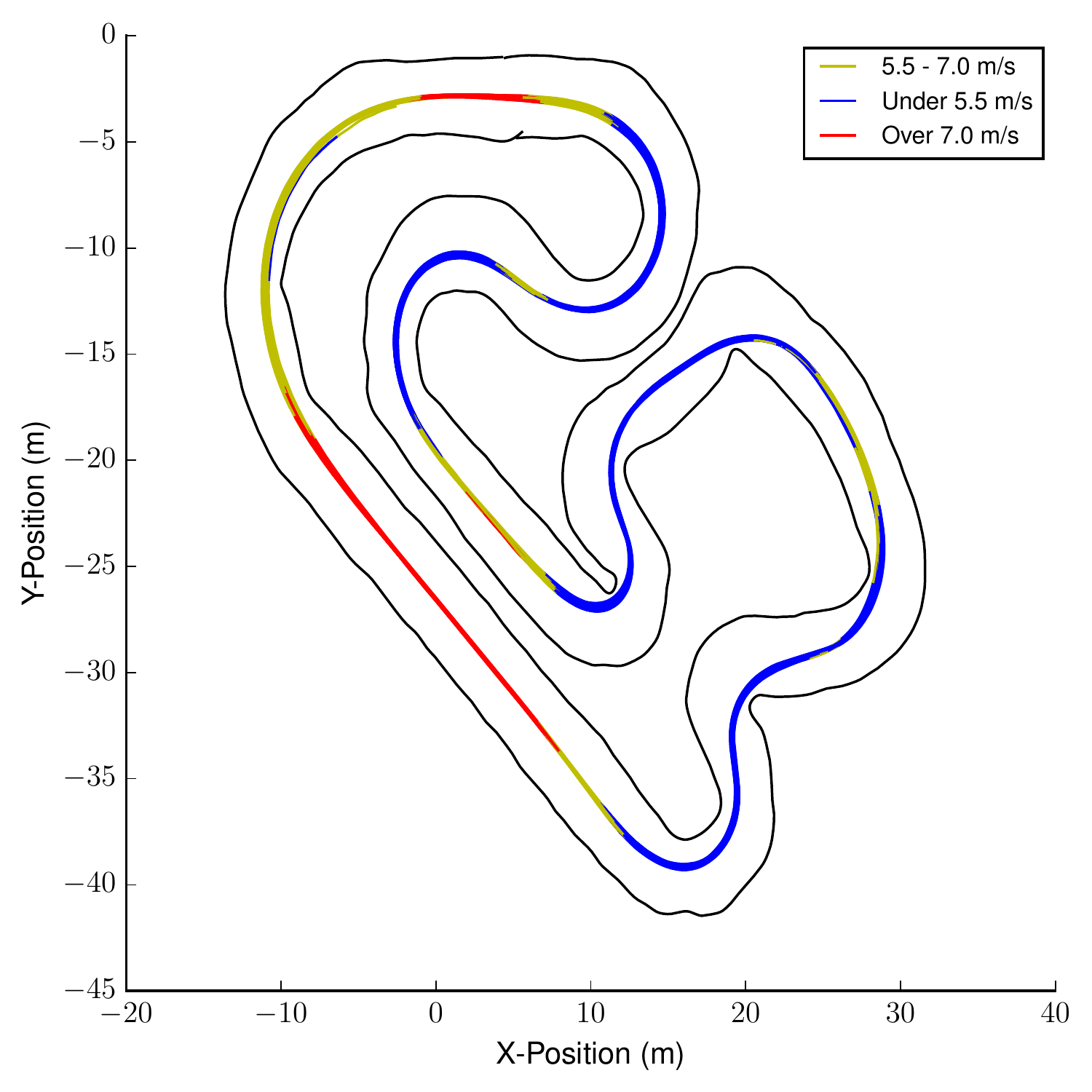}
    \caption{Trajectory traces for simulated autonomous driving runs with model adaptation. Top: Base Model, Bottom: Updated model with LW-PR$^2$. 
    Notice the increased areas of high speed for the LW-PR$^2$ setting compared with the base model.}
    \label{fig:ModelAdaptationTraces}
\end{figure}

\begin{figure}
    \centering
    \includegraphics[width=0.49\columnwidth]{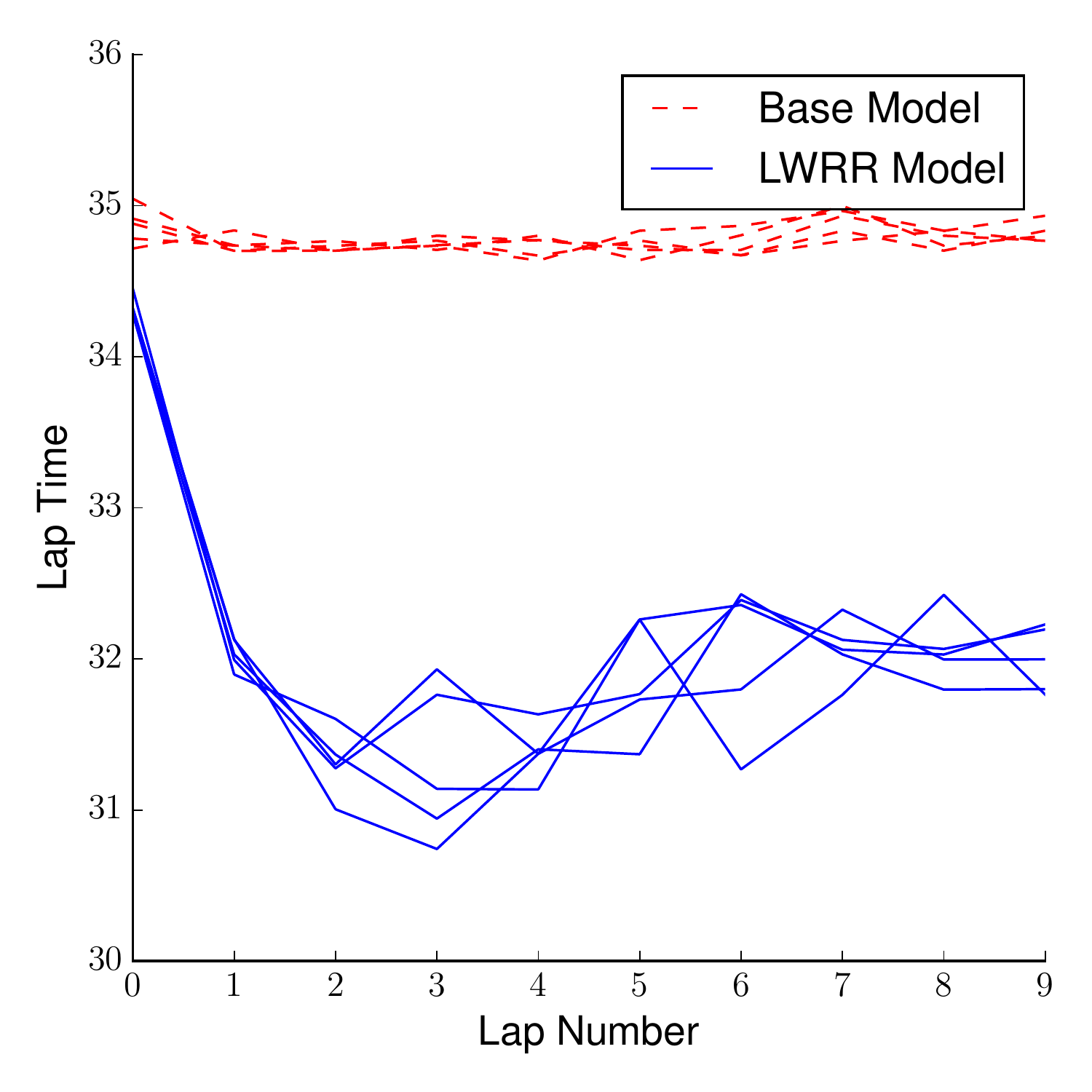}
    \includegraphics[width=0.49\columnwidth]{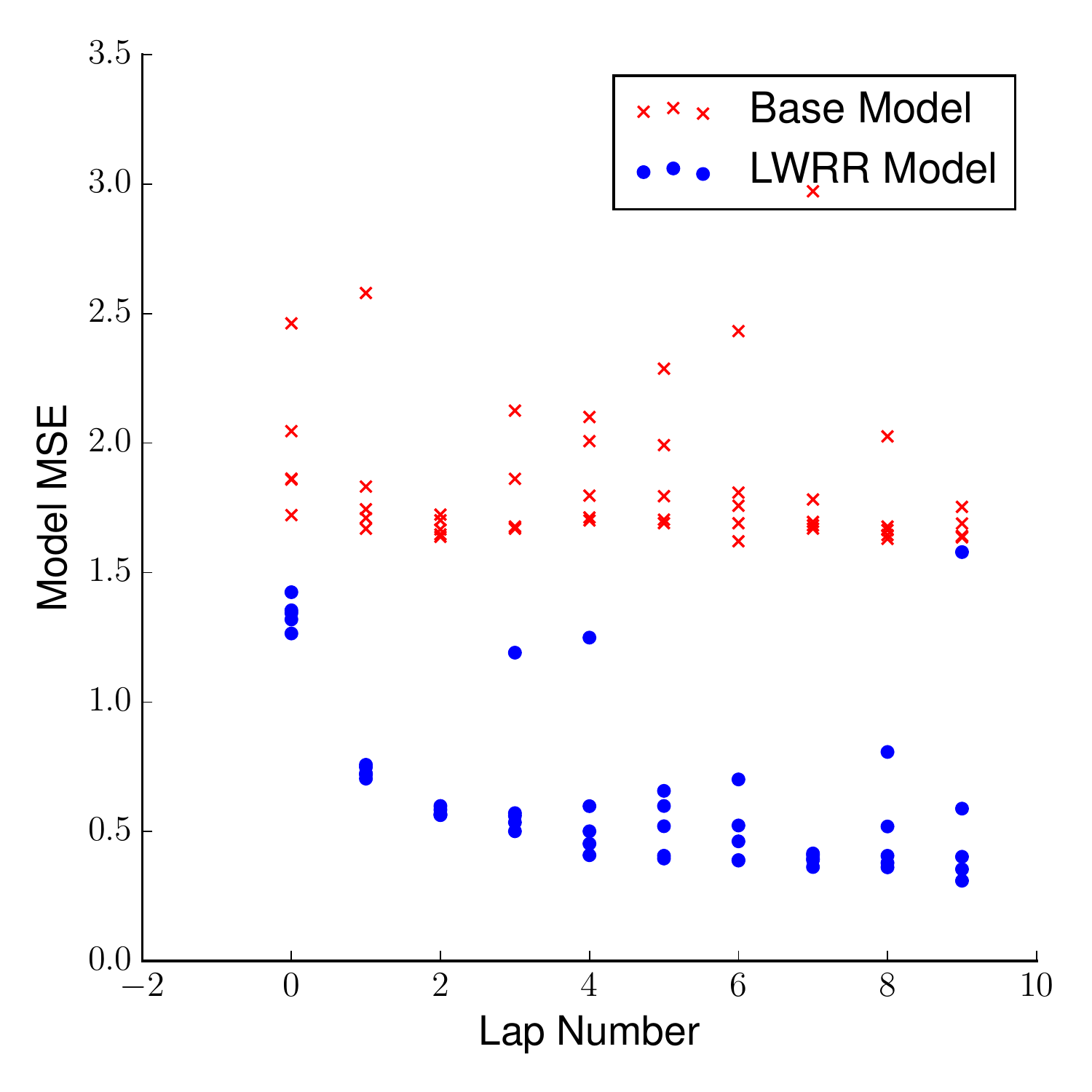}
    \caption{Improvement in lap time and total MSE accumulated per lap for the updated LW-PR$^2$ model versus the base model.}
    \label{fig:LapTime}
\end{figure}

Trajectory traces for the base model and LW-PR$^2$ are shown in Fig. \ref{fig:ModelAdaptationTraces}. The base model performs adequately, and the controller is able to consistently drive the vehicle around the track using the base model. However, the controller with the updated LW-PR$^2$ model is able to attain a higher average velocity around the track.
Figure \ref{fig:LapTime} shows the progression of lap times and total MSE per lap as each trial progresses. On average, it takes less than one lap for the MPC controller to start benefiting from the model adaptation: as the model adapts it realizes it can go faster without slipping in the simulation than it can in the real world and the result is a performance increase. The performance on the second lap is significantly better with the LW-PR$^2$ adapted model than with the base model. After the second lap, the model continues to make small improvements in the per lap MSE.

\subsection{AutoRally Experimental Results}

In the previous sections we tested our model adaptation approach in three essential areas: robustness to catastrophic interference, ability to adapt to drastic changes in the dynamics, and effectiveness when utilized by an MPC controller. These experiments were conducted in a controlled manner using either simulation or specially collected datasets. In this section, we examine how the model adaptation scheme works in a more natural environment - the model adaptation is turned on at the beginning of a day of testing and allowed to run uninterrupted\footnote{Technically the actual program running the model adaptation is interrupted when there are long pauses in testing. However it saves the current parameters, and reloads them when testing resumes.} for the entire day. The resulting dataset consists of over 1 hour of autonomous data collected over a period of 4.5 hours. The 1 hour of autonomous data consists of approximately 18 kilometers of driving data with speeds up to 50 kph. Note that this dataset contains a significant amount of natural variation: the early morning runs are with a fresh damp track, whereas test runs in the afternoon are with a drier track that has less grip. Many of the tests start when the vehicle has a fully charged battery, and then end with a dead battery, and then are continued with another fully charged battery, this means that the adaptation has to constantly re-learn similar parts of the dynamics over again. The tires also become gradually worn out, which has a significant effect on the friction available.

For this dataset we record the same performance metrics as in the earlier online dataset experiments for each of the four adaptation strategies. Additionally, we have available the active performance of LW-PR$^2$ data since the model produced by LW-PR$^2$ was being used to drive the vehicle autonomously. For the active version of LW-PR$^2$ we used a slightly more conservative learning rate, which explains the performance difference between the active and non-active LW-PR$^2$-SGD algorithms. Note that the errors produced by the vehicle running autonomously in these experiments are on average higher than the errors reported in the previous sections. This is due to a combination of speed and control style: moving faster produces higher accelerations which lead to higher errors than in the catastrophic forgetting test, and the autonomous control system does not produce as smooth of control inputs as the expert human which can also lead to high accelerations relative to the expert. 

The results over the full day of testing are given in Table \ref{Table:FieldTestPerformance}. Once again, all of the incremental method significantly improve the performance from the base model. The SGD and LW-PR$^2$ methods perform nearly identically on the online test, but the method updated with LW-PR$^2$ is able to be safely utilized by an MPC controller. 

\begin{table}[htb!]
\caption{Autonomous Field Test}
\centering
\begin{tabular}{c|c|c|c|c} 
                            & Base & SGD & LW-PR$^2$ & LWPR  \\ \hline
Roll Rate ($rad./s$)        &   0.01       &0.01  &0.01         & 0.01              \\ \hline
Longitudinal Acc. $(m/s^2)$ &   2.73       &2.28  &2.30         & {\bf{2.06}}             \\ \hline
Lateral Acc. $(m/s^2)$      &   1.71       &1.29  &{\bf{1.24}}         & 1.28              \\ \hline
Heading Acc. $(rad./s^2)$   &   8.28       &{\bf{4.48}}  &4.87         & 4.54             \\ \hline
Total MSE                   &   3.18       &2.10  &2.11         & {\bf{1.97}}              \\ \hline \hline
Active Performance (MSE)    &      N/A     & N/A  & 2.54        & N/A
\end{tabular}
\label{Table:FieldTestPerformance}
\end{table}

\section{Conclusion} 
\label{sec:conclusion}

In this paper we have presented a novel method for pseudo-rehearsal that is applicable to learning vehicle models in changing environments. The key contributions are: 
\begin{enumerate}
\item Using an incrementally updated LWPR model in order to create artificial training pairs. The use of incrementally updated LWPR enables pseudo-rehearsal to be applied to systems where both the input and target distributions are non-stationary.
\item Using a constrained gradient update, which ensures that the adaptation cannot move away from the system identification dataset, no matter how large the errors it encounters in other regions of the state-space are.
\end{enumerate}
In order to test our method we created a series of datasets and simulation tests that stressed essential requirements for an online adaptation scheme: the ability to prevent catastrophic forgetting, adapt to drastic changes in the dynamics, and the ability to produce models usable by an MPC controller. These experiments demonstrated the capability of our approach and also highlighted some of the nuances involved in validating online adaptation algorithms: in all of the online experiments the standard SGD method performed at the level of the LWPR and LW-PR$^2$ algorithms, it was not until the SGD method was tested on a specifically collected validation dataset or tried to be combined with an MPC controller, that the deficiencies of the SGD method became apparent. We have also demonstrated the practicality of the approach by performing an extended test of the method, where it is required to run continually for hours at a time while producing models that are capable of high speed driving. 


\bibliographystyle{plainnat}
\bibliography{references}

\end{document}